\pdfoutput=1

\documentclass[11pt]{article}
\usepackage{gb4e}
\noautomath
\usepackage[]{emnlp2021}

\usepackage{times}
\usepackage{latexsym}
\usepackage{array}
\usepackage{amsmath}
\usepackage{cleveref}
\usepackage[T1]{fontenc}

\usepackage[utf8]{inputenc}

\usepackage{microtype}
\usepackage{times}
\usepackage[shortlabels]{enumitem}
\usepackage{booktabs}
\usepackage{latexsym}
\usepackage{amssymb}

\usepackage{caption}
\usepackage{subcaption}

\usepackage{float}
\usepackage{microtype}
\usepackage{graphicx}
\usepackage{tikz-dependency}

\title{The Language Model Understood the \textit{Prompt} was Ambiguous: \\
Probing Syntactic Uncertainty Through Generation}

\author{Laura Aina \\
  Universitat Pompeu Fabra \\
  Barcelona, Spain \\
  \texttt{laura.aina@upf.edu} \\\And
  Tal Linzen \\
  New York University \\
  New York, NY \\
  \texttt{linzen@nyu.edu} \\}

\begin{document}
\maketitle
\begin{abstract}
Temporary syntactic ambiguities arise when the beginning of a sentence is compatible with multiple syntactic analyses.
We inspect to which extent neural language models (LMs) exhibit uncertainty over such analyses when processing temporarily ambiguous inputs, and how that uncertainty is modulated by disambiguating cues. 
We probe the LM's expectations by generating from it: we use stochastic decoding to derive a set of sentence completions, and estimate the probability that the LM assigns to each interpretation based on the  distribution of parses across completions. Unlike scoring-based methods for targeted syntactic evaluation, this technique makes it possible to explore completions that are not hypothesized in advance by the researcher.
We apply this method to study the behavior of two LMs (GPT2 and an LSTM) on three types of temporary ambiguity, using materials from human sentence processing experiments.
We find that LMs can track multiple analyses simultaneously; 
the degree of uncertainty varies across constructions and contexts.
As a response to disambiguating cues, the LMs often select the correct interpretation, but occasional errors point to potential areas of  improvement. 
\end{abstract}

\section{Introduction}

During sentence processing, humans incrementally derive analyses of linguistic expressions. 
At an initial stage, uncertainty regarding the analysis may be present if no contextual cues have been provided to allow for a unique interpretation; this results in a temporary ambiguity~\citep{frazier1978comprehending}.
For instance, the initial sentence portion of (\ref{intro_example}) is ambiguous as to the syntactic function of \textit{band}, which can be a direct object, as in~(\ref{intro_example}a), or an embedded subject as in~(\ref{intro_example}b):
\begin{exe}
\ex The audience knew the band ... \label{intro_example} \begin{enumerate}[a)]
    \item very well. \label{intro_example_a}
    \item was going to come back on stage. \label{intro_example_b}
\end{enumerate}
\end{exe}
\noindent 
Like humans, autoregressive neural language models (henceforth, LMs) need to deal with temporary ambiguities, as they process text incrementally.
In this work, we probe the degree of syntactic uncertainty that LMs maintain when processing temporary ambiguities, using generation (sampling) from those models as an analysis tool.

LMs are trained to output contextual probabilities of words in a next-word prediction task.
In spite of this generic objective, 
these models have been shown to track syntactic information to a remarkable extent~\citep{linzen2020syntactic}
and build context-sensitive internal representations, potentially resolving ambiguities in the input (e.g.,~\citealt{peters2018deep}).
The behavior of LMs on temporary syntactic ambiguities was previously investigated through the lens of word surprisal (e.g.,~\citealt{futrell2018rnns}), providing evidence for incremental syntactic processing in LMs. 
At the same time, the extent that a LM expects each interpretation of an ambiguous input, and therefore its degree of syntactic uncertainty, has not been quantified.

In this paper, we generate text from a LM, as a window into the LM's processing of an unfolding sentence.
As (\ref{intro_example}) shows, the completion of a temporarily ambiguous fragment clarifies the intended interpretation. 
We can use the LM's output probabilities to complete an input -- a \textit{prompt} -- and infer how it was initially interpreted. 
We generate a set of completions by drawing multiple samples from the LM's output distribution.
The proportion of completions that are consistent with a certain parse of the prompt is taken to indicate the degree that that parse is expected by the LM.

We consider three types of temporary ambiguities in English; for each type, we derive prompts from sentences drawn from psycholinguistic experiments~\citep{grodner2003against,frazier1987resolution}.
We compare the LM's uncertainty on ambiguous prompts as well as unambiguous prompts that vary in the number and location of disambiguating cues.
We infer the interpretation of a generated sentence from the labels predicted by a syntactic parser.
The LMs we analyse are the LSTM model released by~\citet{gulordava2018colorless} and the transformer GPT2~\citep{radford2019language}.\footnote{Our code is made available at \url{https://github.com/amore-upf/syntactic-uncertainty-LMs}.}

We find that in the presence of temporary ambiguity LMs can track multiple interpretations in parallel, displaying syntactic uncertainty.
The degree of bias towards one analysis varies across ambiguity types and across specific sentences within each type.
Disambiguating cues often appropriately reduce the LMs' syntactic uncertainty in favor of the correct analysis.
At the same time, we also identify evidence of disambiguation issues in the resolution of NP/Z and Noun/Verb ambiguities.
We complement our analyses with a study on the effect of different decoding strategies on syntactic uncertainty, and a comparison of our method to scoring-based analysis.

\section{Related Work}

Temporary ambiguities have been studied extensively in psycholinguistics, as window into human incremental parsing: in case of ambiguities, is only one or a subset of the possible parses considered~\citep{frazier1978sausage}, or are all parses tracked in parallel, weighted by probability~\citep{hale2001probabilistic}?
After disambiguation, how is the analysis revised~\citep{grodner2003against}, and do traces of initial misinterpretations linger~\cite{christianson2001thematic}?
We consider analogous questions, focusing on LMs instead of humans. 

Several studies have examined the syntactic abilities of LMs, through targeted evaluations on specific syntactic phenomena~(e.g.,~\citealt{linzen2016assessing, wilcox2018rnn}), or by analysing the degree to which syntactic information can be decoded from their internal representations~(e.g.,~\citealt{giulianelli2018under,hewitt2019structural}). 
These studies show that LMs track syntax to a large extent, even when not explicitly trained to do so. 

Some previous studies have investigated the behavior of LMs on temporary ambiguities focusing on the \textit{garden-path effect}~\citep{bever1970cognitive}, where a high cognitive cost at disambiguation is taken to signal a preference for the alternative analysis. 
On the one hand, LMs' next-word probabilities can be used to model these effects~\citep{van2018modeling}.
On the other, one can test whether LMs themselves exhibit garden-path effects, looking at their surprisal at the disambiguation of the sentence~\citep{futrell2019neural}. 
This was taken to indicate that, as the sentence unfolds, the LM maintains a representation of its syntactic state, akin to an incremental parser.

As we do in the present work, \citet{futrell2018rnns} generated completions from language models, but that study focused on LMs' awareness of obligatory syntactic events (specifically in the case of relative clause completions).
Finally, \citet{van2021single} analyzed the syntactic predictions of LMs after temporary ambiguous sentence portions, but limited their analysis to the part-of-speech probabilities for the single word that is expected to follow the ambiguous portion. 
Although it is limited to one word, this approach is related to ours as the expectations of the LM are analyzed classifying its predictions based on syntactic information.

\section{Temporary Ambiguities}
\label{sec:ambiguities}

This section describes the types of temporary ambiguity and the materials we use in our study~(the full list of materials can be found in Appendix \ref{appendix_d}).
For each type of ambiguity, we refer to the ambiguous word whose syntactic role determines the analysis of the sentence as the \textbf{locus of the ambiguity}.

\subsection{The NP/S Ambiguity}
The sentence portion in~(\ref{nps}) is compatible with the main verb \textit{understood} taking either a noun phrase~(NP) or a sentential~(S) complement.
This is reflected by the syntactic role of \textit{contract} -- the locus of the NP/S ambiguity -- which could act as direct object of the main verb~(\ref{nps}a), or as embedded subject in an upcoming subordinate clause~(\ref{nps}b). 

\begin{exe}
\ex The employees understood the contract ... \label{nps}
\end{exe}\vspace{-.1cm}
2a) \underline{NP}: \vspace{-.4cm} \\ 
\begin{dependency}[theme = simple]
  \begin{deptext}
  The employees \& understood \& the contract \& well. \\
  \end{deptext}
  \depedge[arc angle = 30]{2}{3}{DOBJ}
  \deproot[edge unit distance=.8ex]{2}{ROOT}
\end{dependency} \\
2b) \underline{S}: \vspace{-.6cm} \\ 
\begin{dependency}[theme = simple]
\begin{deptext}
\hspace{-1.1cm}The employees \& \hspace{-1.1cm}understood \& the contract \& would  \\ be changed very soon. \\
  \end{deptext}
  \depedge[arc angle = 20]{4}{3}{NSUBJ}
  \depedge[arc angle = 40]{2}{4}{CCOMP}
  \deproot[edge unit distance=1.1ex]{2}{ROOT}
\end{dependency}

\noindent An equivalent of~(\ref{nps}b) without temporary ambiguity can be obtained by adding the complementizer \textit{that}:
\vspace{0.2cm}

\noindent 2c) \underline{S}: The employees understood that the contract would be changed very soon. \label{nps-pair} \vspace{0.2cm}

We use the 20 NP/S sentence pairs from~\citet{grodner2003against}.\footnote{\citet{grodner2003against} considered two variants of sentences, with or without material between the locus of ambiguity and the post-locus cue~(\textit{modified} and \textit{unmodified}, respectively). We only use the unmodified sentence pairs for both NP/S and NP/Z.\label{fn:grodner}}
Each pair consists of a temporarily ambiguous sentence and its unambiguous counterpart, both of which eventually have an S interpretation~(\ref{nps}b and~\ref{nps}c).
From each sentence pair, we derive four types of \textbf{prompts}; i.e., the sentence portion which is passed to the LM as input for generation.
Examples of prompts are shown in Table~\ref{prompts_nps}.
\textsc{No cue} prompts are the sentence portions that exhibit temporary ambiguity.
The other prompts contain at least one disambiguating cue, before or after the locus of ambiguity: \textit{that} is the \textbf{pre-locus cue}, while the \textbf{post-locus cue} is the word immediately after the locus of ambiguity~(varying across items).

\begin{table}[h]
\centering
\small{
\setlength{\tabcolsep}{1pt}
\begin{tabular}{p{4cm}p{3.2cm}} 
Prompt type & \\ \midrule
\textsc{\textbf{no cue}} & The employees knew the  \underline{contract} \\ 
\textsc{\textbf{post-locus cue}} & The employees knew the \underline{contract} \textbf{would} \\
\textsc{\textbf{pre-locus cue}} & The employees knew \textbf{that} the \underline{contract} \\
\textsc{\textbf{pre\&post-locus cues}} & The employees knew \textbf{that} the  \underline{contract} \textbf{would}  \\ \bottomrule 
\end{tabular}
\caption{Examples of prompt types for NP/S; locus of ambiguity underlined, cues in bold. }
\label{prompts_nps}}
\end{table}

\subsection{The NP/Z Ambiguity}
In~(\ref{npz}), the verb \textit{left}  in the subordinate clause can be parsed as taking either a noun phrase complement~(NP) or none~(zero complement; Z). 
The locus of ambiguity is \textit{party}, which can be direct object of \textit{left}~(\ref{npz}a) or subject of the upcoming main verb~(\ref{npz}b).

\begin{exe}
\ex Even though the band left the party ... \label{npz} \vspace{-0.2cm}
\end{exe}
3a) \underline{NP}: \vspace{-.4cm}\\
 \begin{dependency}[theme = simple]
  \begin{deptext} 
 \vspace{-.2cm} Even though the band \& left \& the party \&  I \& stayed. \\ 
  \end{deptext}
  \depedge[arc angle = 30]{2}{3}{DOBJ}
  \depedge[arc angle = 80]{5}{2}{ADVCL}
  \depedge[arc angle = 30]{5}{4}{NSUBJ}
  \deproot[edge unit distance=1ex]{5}{ROOT}
\end{dependency}\\
\begin{minipage}{0pt}
3b)~\underline{Z}:\\ 
\begin{dependency}[theme = simple]
  \begin{deptext}
Even though the band \& left \& the party \& went \& on \\ \hspace{-1cm} for another hour. \\
  \end{deptext}
  \depedge[arc angle = 70]{4}{2}{ADVCL}
  \depedge[arc angle = 30]{4}{3}{NSUBJ}
  \deproot[edge unit distance=1.2ex]{4}{ROOT}
\end{dependency}
\end{minipage}

\noindent The unambiguous version of~(\ref{npz}b) adds a comma between the subordinate and main clauses:
\vspace{0.2cm}

\noindent 3c) \underline{Z}: Even though the band left, the party went on for another hour. \label{npz-pair} \vspace{0.2cm}

We use the 20 ``unmodified''~(see \Cref{fn:grodner}) NP/Z sentence pairs from~\citet{grodner2003against}.
Both sentences in each pair ultimately had the Z interpretation~(\ref{npz}b and~\ref{npz}c). From a sentence pair, we derive prompts following the same criteria described for NP/S ambiguity; here, the pre-locus cue is the comma.

\subsection{The Noun/Verb Ambiguity}
The last ambiguity we investigate concerns words that can function as either a noun or a verb. Such words can lead to temporary structural ambiguities as in~(\ref{nounverb}): if \textit{suit}, the locus of this ambiguity, is a noun, \textit{pants} acts as its modifier; otherwise, it serves as its subject. 

\begin{exe}
\ex Mary thinks that the pants suit ... \label{nounverb}
\end{exe}
4a) \underline{Noun} \\ 
 \begin{dependency}[theme = simple]
  \begin{deptext}
Mary thinks that \& the \& pants \& suit \& is \& pretty. \\
  \end{deptext}
  \depedge[arc angle = 30]{5}{4}{NSUBJ}
  \depedge[arc angle = 30]{4}{3}{NN}
  \depedge[arc angle = 80]{4}{2}{DET}
\end{dependency} \\
4b) \underline{Verb}\\ 
\begin{dependency}[theme = simple]
  \begin{deptext}
Mary thinks that \& the \& pants \& suit \& me well. \\
  \end{deptext}
  \depedge[arc angle = 90]{4}{2}{DET}
  \depedge[arc angle = 30]{4}{3}{NSUBJ}
\end{dependency}

\noindent The temporary ambiguity in~(\ref{nounverb}) can be preempted by replacing \textit{the} with the determiners \emph{this} and \emph{these}, which, through number agreement, favor one of the interpretations:

\vspace{0.2cm}
\noindent 4c) \underline{Noun}: Mary thinks that this pants suit is pretty.\\
\noindent 4d) \underline{Verb}: Mary thinks that these pants suit me well.\\ 

We study this type of ambiguity using the data from Experiments~1 and~2 of~\citet{frazier1987resolution}. 
For each temporary ambiguity, two sentence pairs are provided, one with a Noun interpretation and one with a Verb interpretation~(\ref{nounverb}a and~\ref{nounverb}c for Noun, and~\ref{nounverb}b and~\ref{nounverb}d for Verb).
A minority of the sentences used by \citeauthor{frazier1987resolution} were disambiguated by cues other than agreement; in our analyses, we discard those items and focus only on examples disambiguated by agreement.  
This leaves us with 26 sentence pairs~(out of 32) each for Noun and Verb interpretations. 
We obtain prompts from the pairs, treating the determiner as the pre-locus cue.
As we have one pair for each reading, for unambiguous prompt types~(all but \textsc{No cue}), we derive two prompt subtypes, one for the Noun reading and another for the Verb reading.

\section{Methods}

\paragraph{Language Models}

We evaluate two English language models: 
the \textbf{LSTM} model from~\citet{gulordava2018colorless}, which was trained on a 80M-tokens Wikipedia corpus, with 2 hidden layers of 650 units; and the transformer-based \textbf{GPT2} (\textit{small};~\citealt{radford2019language}), which was trained on the 40GB WebText corpus, with 12 hidden layers of 768 units.\footnote{GPT2 is used through the \href{https://github.com/huggingface/transformers}{Transformers} library~\citep{wolf-etal-2020-transformers}. For text generation, we adapt the code of the available decoding functions to also work with the LSTM.}
Both LMs are unidirectional: their predictions solely depend on the previous context. 
In the LSTM, this is achieved through recurrent connections, while in GPT2 through masked self-attention. 
Given previous evaluations, more fluent generation can be expected from GPT2, which surpasses the LSTM in both number of parameters and size of training corpus.

\paragraph{Generation}

Starting from a prompt, we generate a completion through stochastic decoding, sampling words from the LM's output distribution.
The LM processes the prompt as input and outputs a probability distribution over the next token (\ref{lm1}); we sample a word from this distribution sampled (\ref{lm2}) and use this word as the next input token.
The process is repeated to generate the next tokens.
\begin{align}
P(X_{i+1} | x_{1:i}) = LM(x_{1:i})  \label{lm1} \\
x_{i+1} \backsim P(X_{i+1} | x_{1:i})\label{lm2}
\end{align}
\noindent To obtain a sentence completion, we generate a fixed number of tokens from the prompt and crop the text to sentence boundaries identified using the Spacy Sentencizer.\footnote{\url{https://spacy.io/api/sentencizer}}
More details are provided in Appendix \ref{appendix_a}.

In our main experiments, we do not apply techniques that modify the LM's output distribution before sampling (e.g., nucleus sampling;~\citealt{holtzman2019curious}).
In Section~\ref{sec:sampling} we analyze how such decoding strategies affect syntactic uncertainty.

\paragraph{Syntactic Uncertainty Estimation}

We consider a scenario where the locus of ambiguity can be interpreted in one of two ways, $i_1$ or $i_2$. 
We aim to estimate the probability that the LM assigns to each interpretation based on the prompt; that is, 
a Bernoulli distribution, where $
P (i_1| \text{prompt}) = 1 - P(i_2 |\text{prompt} )$.
We derive an empirical estimate of this distribution by independently sampling a set of sentence completions of the prompt ($C_\text{p}$) and mapping them to their interpretations. 
We generate 100 completions of each prompt, sampled with replacement; in practice, the same completion is rarely generated more than once. Appendix \ref{appendix_b} reports an analysis of the diversity of completions within a sample, in terms of lexical overlap and the proportion of unique sentences.
The relative frequency of interpretations in the sample is then used to estimate their probabilities:
\begin{align}
\small{
\hat{P}(i_1 | \text{p}) = \frac{|\{c \in C_\text{p}| \text{interpretation}(c) = i_1\}|}{|C_\text{p}|}} 
\end{align}
This allows us to quantify the degree of preference of the LM for each interpretation of the prompt, and thus its uncertainty.
For unambiguous prompts, it would be desirable for the probability of the correct interpretation to be 1, as the LM should only generate completions that are consistent with the correct interpretation.
In the presence of ambiguity (\textsc{No cue} prompts), an LM that implicitly implemented a fully parallel parser~\cite{hale2001probabilistic} would distribute the probability mass across multiple interpretations.

\paragraph{Completion Classification}
This method requires us to classify completions based on the syntactic interpretation they imply for the locus of ambiguity.
Manual classification is highly reliable, but less practical when a large set of sentences needs to be analyzed (in our case, at least 8K per ambiguity type).
By contrast, automatic classification relies on the use of a syntactic parser, which may introduce noise in case the parser itself incorrectly disambiguates the sentence. 

As a compromise, we use automatic annotations, and assess their quality by comparing them to manual annotations for a subset of sentences.  
We use the AllenNLP~\cite{gardner2018allennlp} dependency parser, based on the model of~\citet{dozat2016deep} (label attachment accuracy with predicted PoS tags $=92.86\%$). 
For each ambiguity type we use a set of rules to classify completions based on the predicted labels.
These are summarized in the next few sections, and described in more detail in Appendix \ref{appendix_c}. 
If a completion cannot be traced back to either of the candidate interpretations, it is discarded from the sample; this rarely happens in practice.

For each type of ambiguity, a random sample of 80 sentences (20 for each prompt type) generated by GPT2 is manually annotated.
This is carried out by three trained linguists, each of whom reviews data from a different ambiguity type.
There are four possible labels for each sentence: the two candidate interpretations, as well as \textit{other} if the sentence has a different interpretation than the two candidate interpretation and \textit{unclear} if the sentence cannot be interpreted. 
The annotators also judge the syntactic well-formedness of the sentences. We do not consider the semantic plausibility of the sentence.
For all ambiguity types, between 61\% and 66\% of the generated sentences are judged to be fully well-formed; an additional fraction of the data (9--19\%) are sentences that could be well-formed if it were not for punctuation errors and character errors.
The lack of grammaticality of a sentence does not typically impair the ability to infer the interpretation of the locus of ambiguity (i.e., the annotators rarely used the labels \textit{unclear}). 
We provide additional details on the annotation process in Appendix \ref{appendix_b}.

\section{The NP/S Ambiguity}

\begin{figure}
\centering
\hspace{-0.4cm}
\includegraphics[width = 8cm]{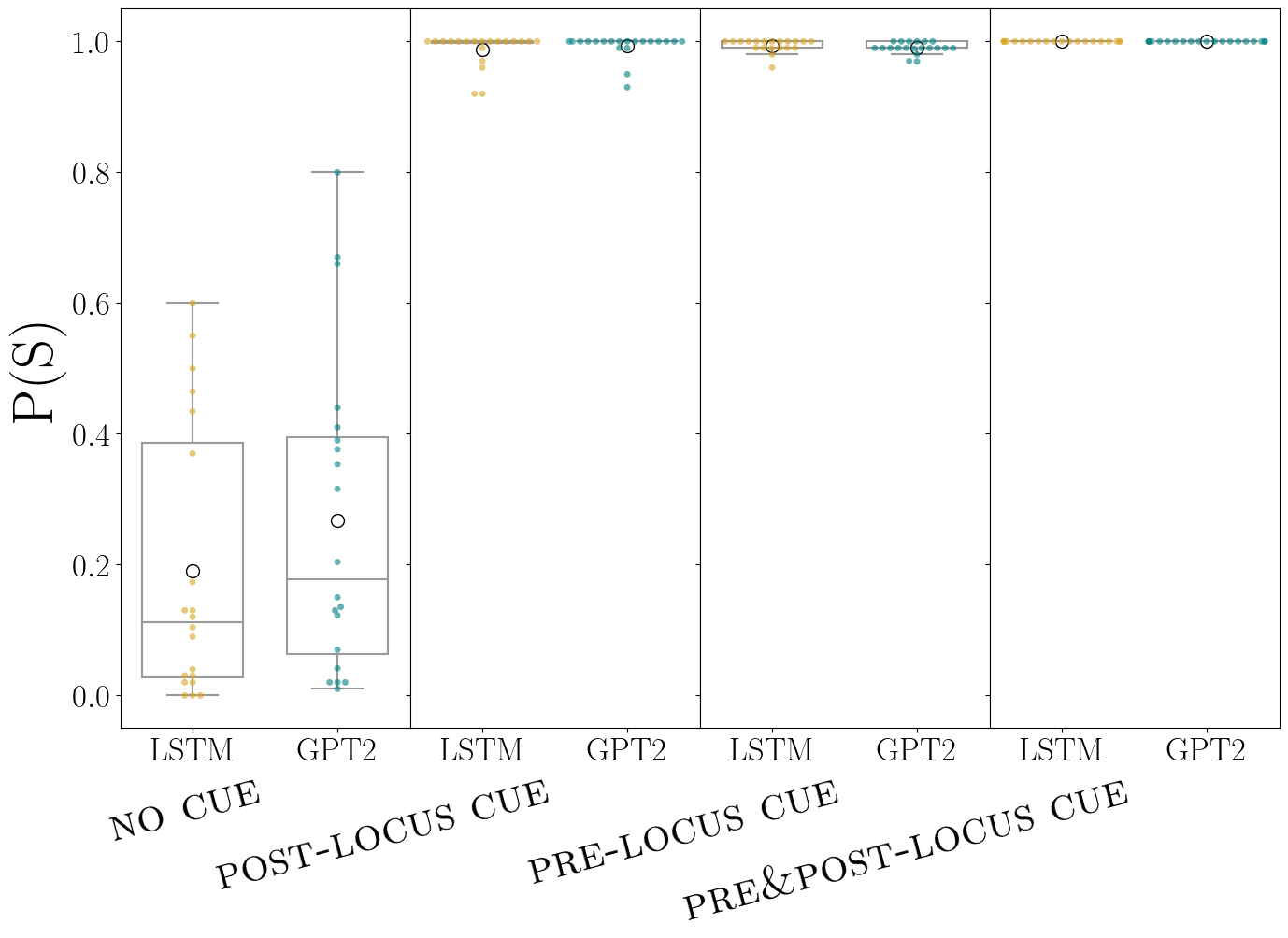} 
    \caption{Distribution of $P(\text{S})$ for each NP/S prompt type and LM; circle = mean across items.}
    \label{fig:np-s}
\end{figure}

\paragraph{Classification} 
To classify a completion as a case of an NP or an S analysis, we inspect the dependency label of the locus of ambiguity (direct object $\rightarrow$ NP; subject $\rightarrow$ S). 
In some of the completions, the locus of ambiguity forms part of a complex NP (e.g., a modifier of another noun, as in \textit{the contract clauses}):
In this case, we use the dependency labels of the following words. 
To reduce noise from parser errors, we define a heuristic that corrects the most typical type of misclassification (NP instead of S).\footnote{The misclassification consists in the locus of ambiguity being labeled as direct object while the finite verb that directly follows it is left without a preceding subject. 
This parse is ungrammatical in English. 
See Appendix \ref{appendix_c_np} for an example and more details.}
If these rules do not identify the interpretation as either NP or S, the sentence is discarded from the completions; for both LMs, this is the case for 0.2\% of the completions, across all prompt types. 
This classification method is reliable:
First, the sentences from which the prompts were derived are all correctly classified as S. 
Second, there is near-perfect agreement between the manual and automatic annotations (Cohen's $k$ = .96; accuracy of automatic classification with respect to manual: .99). 

\paragraph{Results}
Based on the distribution of NP and S completions, we compute $P(\text{S})$ for each prompt ($P(\text{NP}) = 1 - P(\text{S})$). 
The distribution across items for the different prompt types is shown in Figure~\ref{fig:np-s}, and examples of completions can be found in Table~\ref{nps-examples}.

We focus first on the \textsc{No Cue} prompts, which are ambiguous between NP and S. 
The LMs are often uncertain -- to varying degrees -- between the two interpretations. In most cases, they exhibit a preference for NP ($P(\text{S}) < .5$), though this preference is typically not absolute, as S completions are also generated (e.g., (1) in Table~\ref{nps-examples}). 
\begin{table}
    \centering
    \small
    \begin{tabular}{l} 
        \toprule
        (1) The scientist proved the theory \\
        a) \textit{through two experiments.} (NP) \hspace{0.1cm}  b) \textit{was correct.} (S) \\ \midrule
        (2) The tourists saw the palace was \\
        a) \textit{on fire.} (S) \hspace{0.2cm} b) \textit{under construction.} (S)  \\ \midrule
        (3) The journalist confirmed that the story \\
a)\textit{ is false.} (S) b) \textit{was being reported on his network.} (S)  \\ \bottomrule
    \end{tabular}
    \caption{Examples of completions generated by GPT2 for NP/S prompts.\vspace{-0.1cm}}
    \label{nps-examples}
\end{table}
This indicates that, in the presence of the NP/S ambiguity, the LMs tend to consider multiple parses at the same time.
In spite of the general preference for NP, $P(\text{S})$ vary across items, with some cases even favoring an S analysis. 

The other prompt types all contain at least one cue disambiguating the sentence as S; as such, we expect the LM to generate only completions that are consistent with S. 
In line with this prediction, for all these conditions and LMs, $P(\text{S})$ is very close to 1.
This indicates that the LMs are sensitive to the disambiguating cues and use them correctly to adapt their interpretation.
A qualitative inspection of the sentences supports this observation: there is no evidence of disambiguation issues (e.g., (2-3) in Table~\ref{nps-examples}).
\noindent A minority of completions of unambiguous prompts are recognized as NP. 
This occurs due to misclassifications, but also ill-formed completions whose interpretation is unclear (e.g., ``The employees understood that \textit{the contract.}''), or when NP is licensed despite the post-locus cue (e.g., ``The army found the supplies saved \textit{by the French.}'').

\section{The NP/Z Ambiguity}

\paragraph{Classification} As in the case of the NP/S ambiguity, the NP and Z interpretations can be distinguished based on the syntactic role (direct object or subject) of the locus of ambiguity.
We therefore employ the same set of rules we used for NP/S.
The rule correcting cases where a subject is labeled as direct object turns out to be crucial for this classification, as the parser is prone to errors on NP/Z sentences.
A total of 0.6\% of of GPT2 completions and 1.4\%  of LSTM completions cannot be identified as either NP or Z, and are thus discarded from analysis. 
The agreement between the automatic and manual annotations is high (Cohen's $k = .86$; accuracy of automatic classification with respect to manual: .95); the few divergences are cases where the annotator used the label \textit{unclear}, which is not available to the parser.

\begin{figure}
\centering
\hspace{-0.4cm}
\includegraphics[width = 8cm]{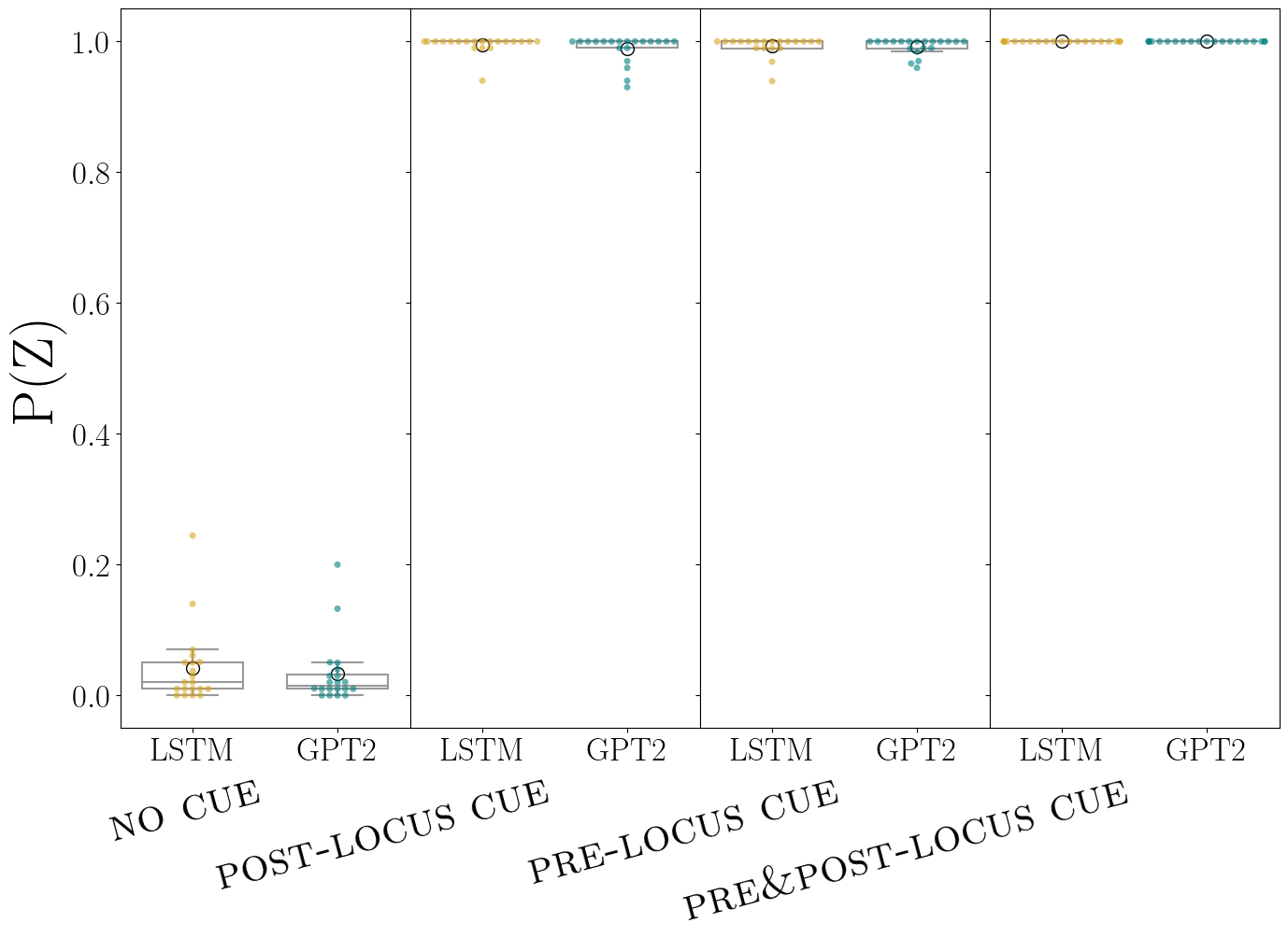} 
    \caption{Distribution of $P(\text{Z})$ for each NP/Z prompt type and LM; circle = mean across items.}
    \label{fig:np-z}
\end{figure}

\begin{table}
    \centering
    \small
    \begin{tabular}{l} 
        \toprule
        (1) In case the executive forgot the assistant \\
        a) \textit{, the assistant was never fired and so on.} (NP) \\
        b) \textit{explains the recommendations to this memo.} (Z)\\ \midrule
        (2) Because the train stopped the traffic was \\
        a) \textit{much slower.} (Z) \hspace{.3cm} b) \textit{suspended immediately. }(Z)  \\ \midrule
        (3) Even though the girl phoned, the instructor \\
a) \textit{ignored her.} (S) \hspace{.3cm} b) \textit{was too rude.} (S)  \\ \bottomrule
    \end{tabular}
    \caption{Examples of completions generated by GPT2 for NP/Z prompts.}
    \label{npz-examples}
\end{table}

\paragraph{Results}
Figure~\ref{fig:np-z} shows $P(\text{Z})$ values ($P(\text{NP}) = 1 - P(\text{Z})$) for each prompt type. 
Examples of completions are reported in Table~\ref{npz-examples}.

In the ambiguous \textsc{No cue} prompts, there is limited syntactic uncertainty:
$P(\text{Z})$ stays close to 0 (on average, .03 and .04 for GPT2 and LSTM), as NP completions are generated much more often than Z ones. 
In spite of this default preference for NP, when there is at least one cue that biases the prompt in favor of the Z reading, $P(\text{Z})$ spikes to 1 or close to it, in line with the expected behavior on unambiguous prompts.
Inspecting the completions, we find that most cases are correctly disambiguated (e.g., (2--3) in  Table~\ref{npz-examples}). 
The handful of NP completions are due to misclassifications or unclear cases, analogously to those reported for NP/S.

Alongside these encouraging results, we also observe the following curious behavior on a subset of completions to \textsc{Post-locus cue} prompts (examples from GPT2):

\begin{exe}
\ex As the couple danced the tango began \textit{, the paparazzi swooned.}\label{paparazzi}
\ex Once the child played the piano was \textit{ours, it was somewhat expected.} \label{piano}
\end{exe}

\noindent 
These completions suggest that even when the Z reading is selected, the LMs may not fully adapt to its structure. We estimate that this behavior affects 8\% and 25\% of \textsc{Post-locus cue} completions of the LSTM and GPT2, respectively (we detect these cases based on patterns in the dependency labels; see Appendix \ref{appendix_c_npz}).
We interpret this phenomenon as evidence of confusion about the structure of the sentence, where the subordinate clause ends up ungrammatically incorporating two predicates. This points to a lingering effect of the initial NP analysis, and difficulty establishing the boundary between the subordinate and main clauses when that boundary is not marked by a comma. 

We note that sentences such as (\ref{paparazzi}) and (\ref{piano}) could, in principle, have a grammatical interpretation if the comma were to be interpreted as conjoining two clauses; under such an interpretation of (\ref{paparazzi}), two things happened during the couple's dance: the tango began and the paparazzi swooned. However, this charitable interpretation is called into question by the fact that such comma-conjoined completions are very rare in other contexts: 0.01\% of all \textsc{Pre\&Post-locus cue} completions, for example, compared to 25\% of \textsc{Post-locus cue} completions.

\section{The Noun/Verb Ambiguity}

\label{sec:nv}

\paragraph{Classification} To classify the generated sentences, we use the PoS label predicted by the parser for the locus of ambiguity.
When run on the sentence pairs that the prompts are derived from, the classification sometimes fails.
To minimize noise, we discard items where the tagger does not correctly interpret at least one of the sentences associated with that ambiguity (4 in total)---we reason that if the parser makes errors on the original sentences, this is likely to occur also on the sentences generated by the LMs from prompts taken from those sentences.
This leaves us with 21 prompts for each subtype.
The agreement between the automatic labels and the annotator's ones is high (Cohen's $k = .83$; accuracy of automatic classification with respect to manual: .91). Differences occur due to tagging errors and sentences annotated as \textit{unclear} by the linguist.

\begin{figure}[t]
\small
\begin{subfigure}{.5\textwidth}
\centering
\includegraphics[width = 7.5cm]{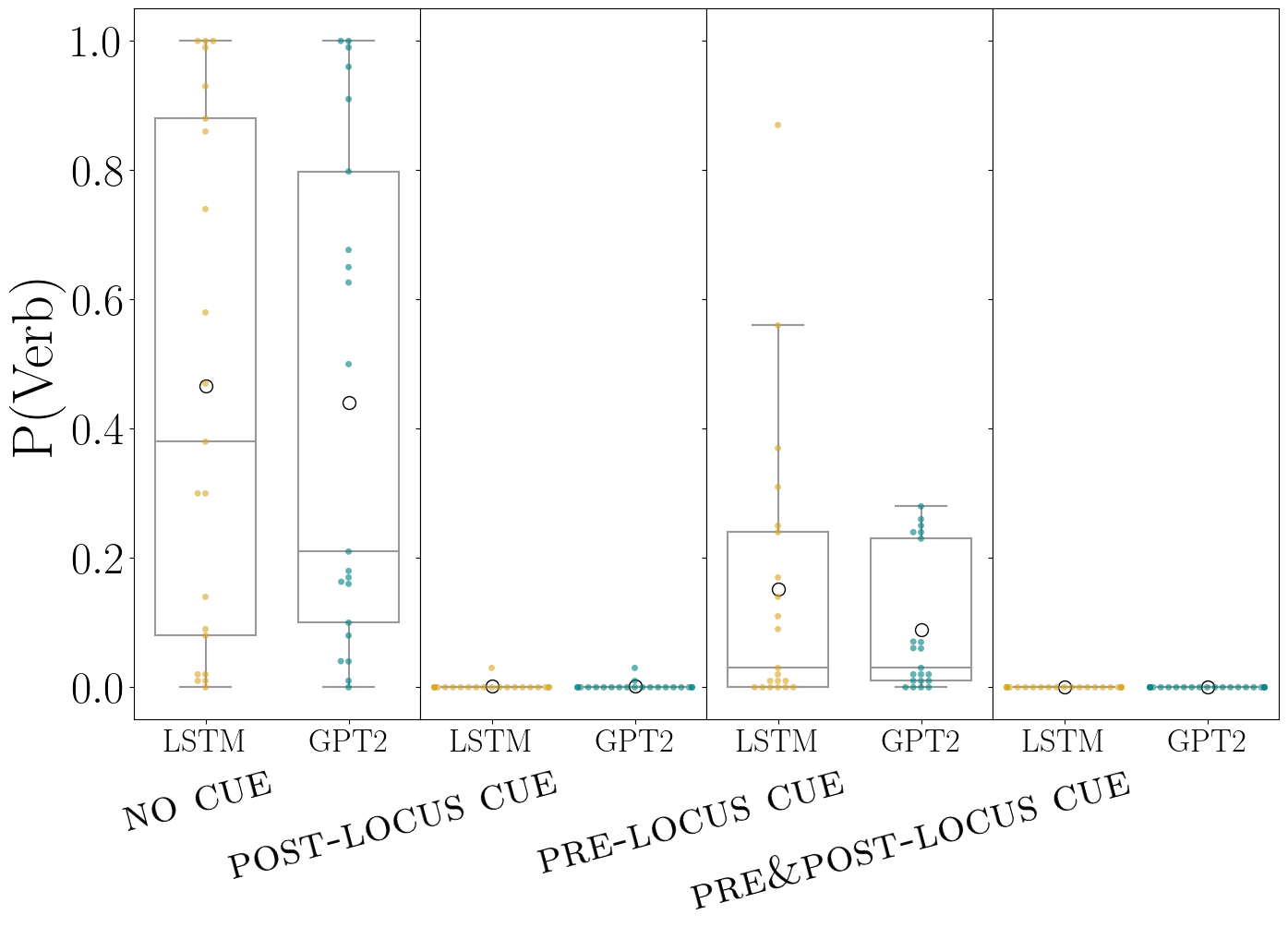} 
\caption{Prompts with Noun interpretation}
\end{subfigure} \vspace{0.2cm} \\
\begin{subfigure}{.5\textwidth}
\centering
\includegraphics[width = 7.5cm]{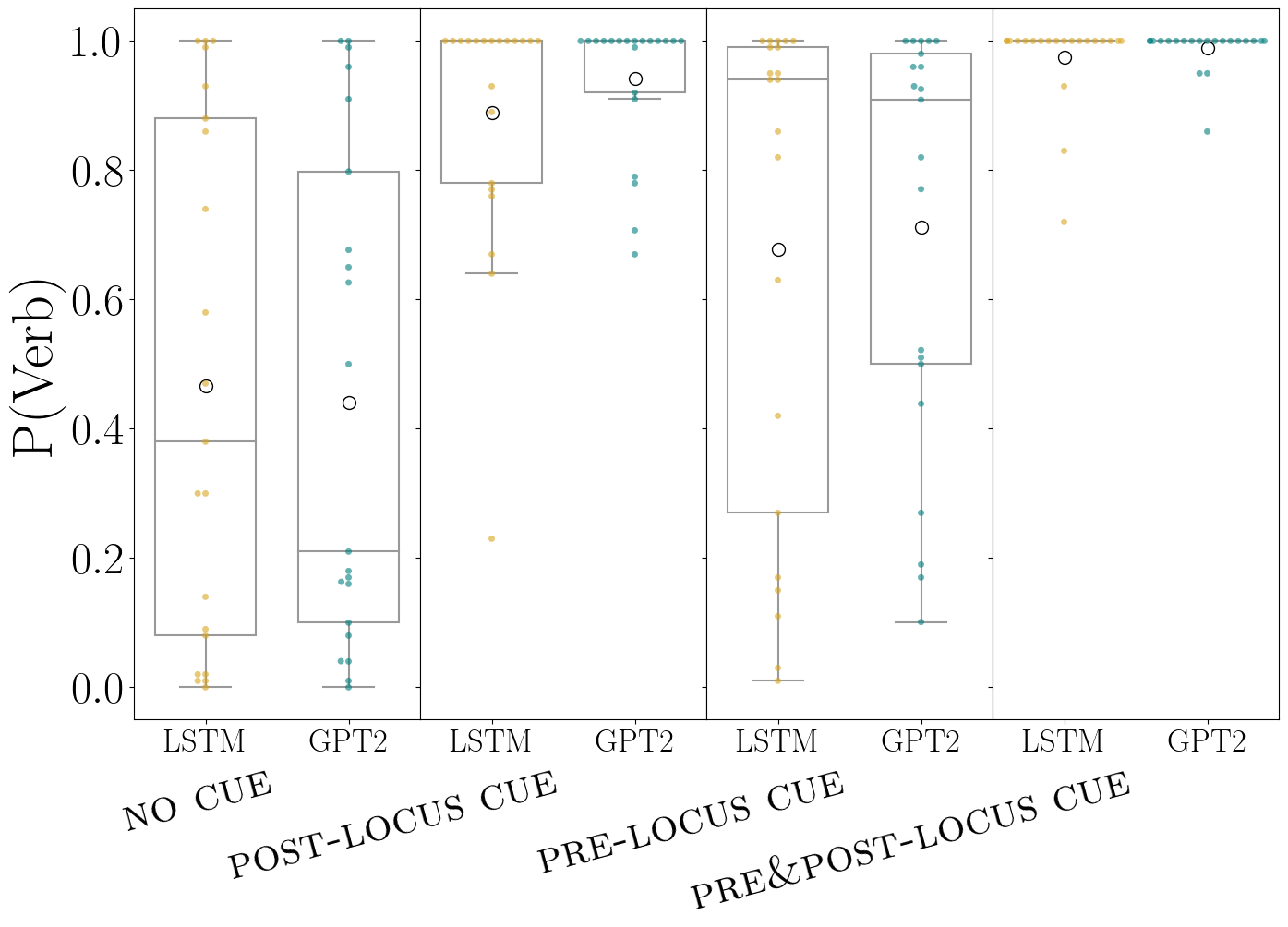} 
\caption{Prompts with Verb interpretation}
\end{subfigure} 
\caption{Distribution of $P(\text{Verb})$ for each Noun/Verb prompt type and LM; circle = mean across items.}
\label{fig:n-v}
\end{figure}

\begin{table}[t]
    \centering
    \small
    \begin{tabular}{l} 
        \toprule
        (1) Nobody knows if it's true that the university fines \\
        a) \textit{are ever issued.} (N) \hspace{.05cm} b) \textit{people who don't study.} (V) \\ \midrule
        (2) Mrs.~Baker is convinced that the school fears are \\ 
        a) \textit{valid points.} (N) \hspace{.2cm} 
        b) \textit{unfounded.} (N)  \\ \midrule
        (3) Mary thinks that the pants suit me \\
a) \textit{better.} (V) \hspace{.2cm}  b) \textit{really in a bad way.} (V)  \\ \midrule
        (4) Despite last year's report, those city hopes  \\
a) \textit{varied.} (N) \hspace{.2cm}  b) \textit{to become wealthier.} (V) \\ \midrule
(5) I know that this desert trains  \\
 a) \textit{people to work!} (V) \hspace{.2cm}   b) \textit{are closed.} (N) \\
\bottomrule
    \end{tabular}
    \caption{Examples of completions generated by GPT2 for Noun/Verb prompts.}
    \label{nv-examples}
\end{table}

\paragraph{Results}$P(\text{Verb})$ values for each prompt type are shown in Figure~\ref{fig:n-v}, while examples of completions can be found in Table~\ref{nv-examples}.

For ambiguous prompts -- i.e.,~\textsc{No cue} -- the dispersion of values is very high for both models. 
Though there is on average a preference for the Noun reading (mean $P(\text{Verb})\approx .4$), the probability assigned to this interpretation varies across items.
This indicates that the LMs' initial syntactic preferences on this temporary ambiguity are highly dependent on its instance.

For \textsc{Post-locus cue} prompts, $P(\text{Verb})$ adapts to the disambiguating cues, approaching 0 for the Noun reading and 1 for the Verb reading.
In the latter case, we find dispersion of values, due to some completions labeled as NP.
A qualitative inspection shows that this occurs due to tagger errors or when a Noun reading is licensed in spite of the post-locus cue (e.g., ``some metal rings loudly \textit{beat into our ears.}'').
Overall, we do not find evidence of disambiguation issues on this type of prompt (e.g., (2-3) in Table~\ref{nv-examples}). 

By contrast, \textsc{Pre-locus cue} prompts, especially in the Verb subtype, pose more challenges to the LMs.
$P(\text{Verb})$ values follow the expected trends -- decreasing for the Noun cases, and increasing for the Verb cases -- but exhibit variation. 
In some Verb cases, we do not even find a preference for the Verb reading (i.e., $P(\text{Verb}) < .5)$. 
\textsc{Pre-locus cue} prompts are disambiguated by the number of the determiner:
These results suggest that the LMs are not fully responsive this cue, especially when it points to a Verb reading.
The LSTM shows greater dispersion than GPT2, indicating greater disambiguation difficulty. 
A qualitative analysis confirms these observations: besides a portion of tagger errors, we find several completions that persist in the incorrect interpretation, violating number agreement (e.g.,~(4b) and (5b) in Table~\ref{nv-examples}).

In general, we find that classification errors occur more often with the Noun/Verb ambiguity than with the previously analysed ones.
As errors introduce noise, our quantitative estimates should be considered approximate.
However, as mentioned earlier, the trends they point to are reliable as they are all confirmed by qualitative analyses of the data.

\section{Effect of Decoding Strategy}
\label{sec:sampling}

In our previous experiments, we generated completions by sampling from the LM's output distribution. 
We compare this approach to other decoding strategies, focusing on NP/S \textsc{No cue} prompts.

\paragraph{Stochastic Decoding} Variants of stochastic decoding modify the LM output distribution before sampling words from it. 
Restricting or biasing the sampling process to high probability words can improve the quality of the generated text~\citep{holtzman2019curious}.
In \textbf{nucleus sampling}, the LM distribution is truncated to top-probability words with cumulative probability $p$~(e.g., $p = 0.9$). 
Another technique modifies the distribution by dividing the output scores by a parameter $t$ -- \textbf{temperature} -- before softmax is applied:
If $t \in [0, 1 )$, the distribution is skewed towards high probability words.

We inspect how these decoding strategies affect the diversity of interpretations of the completions of an ambiguous prompt.
For different combination of values of $p$ and $t$, we generate set of completions from prompts and report the average $P(\text{S})$~(Table~\ref{nps_hyperparameters}).
Standard sampling, used in our previous experiments, corresponds to $p = 1$ and $t=1$.
The values of $P(\text{S})$ decrease as the hyperparameters are modulated to focus on high-probability words~($t$ or $p$ decreases), showing that fewer S completions are generated.
This indicates that temperature and nucleus size can influence how temporary ambiguities manifest in generated text:
Focusing on top-probability words increases the bias our method identifies towards the preferred interpretation~(NP, in the case of NP/S).

\paragraph{Maximization-Based Decoding }
Is the preference for an analysis observed sampling multiple completions reflected by the analysis of the top-probability completion?
We use \textbf{beam search} as our decoding strategy, returning the completion ranking highest in probability~(beam size $= 16$). 
As in this case we consider only one completion per prompt, $P(\text{S})$ for beam search in Table~\ref{nps_hyperparameters} reflects the proportion of prompts whose top-completion has an S interpretation.
Most -- though not all -- have an NP interpretation.
This mirrors what observed with sampling: in NP/S temporary ambiguities, there is a general preference for NP, but this does not apply uniformly to all instances.

\begin{table}
    \centering
    \begin{tabular}{lp{.2cm}p{.2cm}cc}
        & $p$ & $t$ & LSTM & GPT2\\ \toprule
        Pure sampling & 1 & 1 & .19 & .27 \\ \midrule
       Nucleus sampling &  .9 & 1 & .18 & .24  \\
       &  .75 & 1 & .15 & .23  \\
       &  .6 & 1 & .15 & .22  \\ \midrule
       With temperature &  1 & .9 & .18 & .24  \\
       &  1 & .75 & .15 & .24  \\
         & 1 & .6 & .14 & .23 \\ \midrule
       Beam search - 16 & - & - & .10 & .25 \\ \bottomrule 
    \end{tabular}
    \caption{Average $P(\text{S})$ for decoding strategies on \textsc{No Cue} NP/S prompts~($p$: nucleus size; $t$: temperature).}
    \label{nps_hyperparameters}
\end{table}

\section{Comparison to Surprisal-Based Analysis}

Previous work has probed the syntactic state of a LM in a temporarily ambiguous sentence by measuring surprisal (negative log probability) at the disambiguation point~\citep{futrell2019neural}:
If surprisal is higher than in the unambiguous sentence, we can infer that LM initially preferred the alternative, ultimately incorrect analysis.
Focusing on NP/S sentence pairs, we compare this approach to our method to extract probabilities of analyses.

We calculate~(1) the difference in word surprisal of the disambiguating word that follows the locus of ambiguity~(i.e., the post-locus cue) between the ambiguous and unambiguous sentences of a pair; and~(2) the estimated $P(\text{S})$ values on \textsc{No cue} prompts.
For both LMs, we find a strong negative correlation between the difference in surprisal and $P(\text{S})$~(GPT2: Spearman's $\rho =  -.70$; LSTM: $\rho = -.81$; both $p <.05$): As $P(\text{S})$, as estimated through generation, increases, the LM is less surprised that the S analysis is introduced on the temporarily ambiguous sentence than on its unambiguous equivalent.

The fact that the two methods -- surprisal and generation-based -- are aligned in their estimates corroborates the findings derived from either approach.
Both methods can be considered alternatives to probe the syntactic state of a LM, and 
one or the other may be favored depending on the study.
Surprisal estimates can be easily extracted from a LM, making them a more straightforward tool to apply.
However, to analyse the expectations of a LM on an ambiguous input, the surprisal method requires a comparison to its unambiguous counterpart, which is not needed with our method.
This allowed us to study a LM's behavior also on unambiguous inputs, providing insights about the sensitivity to disambiguating cues.
Moreover, generating text exemplifies the LM’s expectations over the sentence, which can reveal phenomena that may not be clear on the basis of surprisal estimates alone. 
An example are the ungrammatical blended continuations of NP/Z prompts~(e.g., "*As the couple danced the tango began\textit{, the paparazzi swooned.}"). 

\section{Discussion and Conclusions}
In this work, we have probed the syntactic uncertainty of LMs by generating sentence completions from the LMs. 
Our results contribute to research on the syntactic processing of LMs, quantifying the extent that one analysis of the input is implicitly entertained by the model. 

We find that when processing temporary syntactic ambiguities, LMs typically exhibit uncertainty about the analysis of the input; that is, they simultaneously consider multiple analyses to be viable.
In line with previous analyses of LMs on garden-path sentences, on NP/S and NP/Z sentences we detected a general preference for the NP analysis (e.g.,~\citealt{van2018modeling}).
But, while for NP/Z this preference was near-absolute and consistent across cases, on NP/S and Noun/Verb ambiguities the LMs' behavior varied across specific instantiations of the ambiguity.
A promising direction for future work would be to determine whether this inter-item variation mirrors human expectations~\cite{ford1982competence,garnsey1997contributions} and/or corpus statistics~(for Noun/Verb, e.g., how often \textit{suit} is used as Noun vs.~Verb).
A plausible hypothesis is that a LM acquires default preferences for analyses from regularities in the training data.

When disambiguating cues are given as part of the input, the LMs tend to display the correct behavior: we observe the appropriate shifts in uncertainty in favor of the disambiguated parse, providing further evidence of their context-sensitivity.
At the same time, certain issues arise for NP/Z and Noun/Verb ambiguities, suggesting that there is room for improvement in the LMs' responsiveness and adaptation to disambiguating cues.
On NP/Z ambiguities, some generated completions exhibit confusion over the sentence structure.
This behavior calls to mind lingering effects of initial misinterpretations found in humans~\citep{christianson2001thematic}.
On Noun/Verb ambiguities, the LMs sometimes failed to use the number of a preceding determiner as cue for the correct parse.
This may be due to difficulties in tracking number agreement in these constructions (in contrast to results by \citet{linzen2016assessing} and subsequent research on subject-verb agreement), or in generally overriding a default preference for the incorrect analysis.

Using generation proved to be an informative tool to inquire a LM's uncertainty over an unfolding sentence, and could be used also to inquire more types of ambiguities~(e.g.,~semantic). 
Yet, there are some challenges to our proposed methodology.
First, relying on an automatic classification of sentences can introduce noise: ambiguities can be difficult for NLP systems even when explicitly trained to analyse expressions~\citep{elkahky2018challenge}.
Second, we could not automatically detect ungrammatical completions or with an unclear analysis~(the parser always returns an output), whereas it may be useful to be identify these cases.
While these issues did not prevent us from inferring the main trends in the LMs behavior, all confirmed by qualitative inspections of the data, we look forward to
future work that will attempt to overcome the aforementioned limitations.

\paragraph{}

\section*{Acknowledgements}
We thank the members of the JHU/NYU Computation and Psycholinguistics Lab and of the UPF Computational Linguistics and Linguistic Theory group for valuable discussions and feedback. 
We are grateful to the annotators for their participation to this research. 
This work was supported in
part by National Science Foundation grant BCS-2020945, and has received funding from the European Research Council (ERC) under the European Union's Horizon 2020 research and innovation programme (grant agreement No. 715154). This paper reflects the authors' view only, and the EU is not responsible for any use that may be made of the information it contains. 

\bibliographystyle{acl_natbib}
\bibliography{main}

\clearpage
\appendix
\section*{Appendices}
\section{Generation: Further Details}
\label{appendix_a}

To generate completions from the prompt, we apply stochastic decoding as described in the paper.
The LSTM employs a word-level encoding, with a fixed vocabulary size, whereas GPT2 uses Byte-Pair-Encoding, with a vocabulary of both word and subword units. 
From a prompt, we generate 30 and 50 tokens for the LSMT and GPT2, respectively.
This is because GPT2 can generate subwords, and thus require more steps on average to reach the end of a sentence.
For the analyses, we discard GPT2 completions where the last word of the prompt is followed by a subword, thus changing its identity (e.g., from \textit{suit} to \textit{suitable}). 
For the LSTM we penalize the generation of the unknown-word symbol, reducing its output score by a factor of $10^{16}$.

A minority of words in NP/Z and Noun/Verb prompts were not in the vocabulary of the LSTM.
We replace these words with equivalent ones that do not substantially affect the meaning of the sentence (e.g., \textit{jogger} $\rightarrow$ \textit{runner}) and use these modified prompts for the experiments on both LMs.

\section{Analysis of Generated Sentences}
\label{appendix_b}

\paragraph{Diversity} We analyse the diversity of the completions generated for each prompt.
Completions were rarely completely identical: the average proportion of unique completions in a sample (pooling together all prompt types) is at least 98\% for all ambiguity types and LMs.
Of course, it is possible for two completions to be very similar, though not identical. To measure to extent of this phenomenon, we measure the lexical overlap across completions, focusing on unigrams and bigrams.
We calculate individual Self-BLEU scores of each completion with respect to the others generated for the same prompt.
Average unigram scores tend to be much higher than the bigram ones across ambiguity types and LMs (the former in the range .68-.71, while the latter .18-.25). 
This shows that individual words are often repeated across completions, but not so frequently in the same order. 

\paragraph{Grammaticality} 
During the manual annotation of the subset of GPT2 completions, the annotators are also requested to provide binary judgments of the syntactic well-formedness of each sentence.
Since character and punctuation errors are frequent (e.g., a misspelled word, or the incorrect presence of a punctuation mark), annotators can specify when a sentence would count as grammatical without such errors. 
Overall, 66\% of NP/S completions, 61\% of NP/Z completions and 66\% of of Noun/Verb completions are judged to be fully well-formed.
If we ignore spelling and punctuation errors, these percentages increase to 75\%, 74\% and 85\%.

\section{Classifying Completions}
\label{appendix_c}

This appendix presents the rules employed to classify completions based on the inferred syntactic interpretation of the prompt. 

\subsection{NP/S and NP/Z Sentences}
\label{appendix_c_np}

It is possible to distinguish between the NP and S interpretations and between the NP and Z interpretation by examining the syntactic role of the head of the noun phrase that contains the locus of ambiguity.
In the simplest case, the locus of ambiguity is the direct object or subject; in other cases, it is part of a complex NP, where, for instance, it modifies another noun. 

\begin{dependency}[theme = simple]
  \begin{deptext}
  \hspace{-.2cm}The employees \&\hspace{-.2cm} understood \&\hspace{-.2cm} the contract \&\hspace{-.2cm} ... \\
  \end{deptext}
  \depedge[arc angle = 40, dotted]{2}{3}{DOBJ}
  \depedge[edge unit distance=1ex, dotted]{3}{4}{NSUBJ}
\end{dependency}
\begin{dependency}[theme = simple]
   \begin{deptext}
   \hspace{-.1cm}The employees \&\hspace{-.2cm} understood \&\hspace{-.2cm} the contract \&\hspace{-.2cm} clauses \&\hspace{-.2cm} ... \\
   \end{deptext}
   \depedge[arc angle = 40, dotted]{2}{4}{DOBJ}
   \depedge[edge unit distance=1ex, dotted]{4}{5}{NSUBJ}
   \depedge[arc angle = 30]{4}{3}{NMOD}
\end{dependency}

We define a set of rules that are based on the dependency labels predicted by the parser.
The rules are applied recursively:
\begin{itemize}
    \item In the base case, we check if the predicted label of a given token is that of direct object or subject. If it is a subject, we return S for NP/S and Z for NP/Z. 
    \item If the label for the given token is neither subject nor direct object, we consider whether it could be part of a complex NP. If the predicted label is that of a modifier or possessive, we apply the function to the following tokens to identify the head of the NP and determine whether it is a subject or direct object. If this scenario does not apply, we return \textit{other}. 
\end{itemize}

The most common parser error involves a failure to detect S or Z cases, labeling the locus as direct object when followed by a finite verb:
\begin{dependency}[theme = simple]
   \begin{deptext}
   \hspace{-0cm} The mechanic \& accepted \& the car \& looked \& great \\
   \end{deptext}
   \depedge[arc angle = 50]{2}{4}{CCOMP}
   \depedge[arc angle = 30]{2}{3}{DOBJ}
   \deproot[edge unit distance=1.3ex]{2}{ROOT}
\end{dependency} 
\noindent This parse is not only incorrect but also ungrammatical, as it leaves the verb after the locus without a subject.
We modify the base case rule to detect and correct these cases:
\begin{itemize}
    \item If the token is a direct object but it is followed by a token with a dependency label compatible with a finite verb (e.g., \textit{root, ccomp}), we change the label to subject, and return S for NP/S and Z for NP/Z.
\end{itemize}

\subsection{NP/Z Sentences with Disambiguation Issues}
\label{appendix_c_npz}

In Section 6, we described a subset of completions to \textsc{Post-locus cue} completions of NP/Z prompts that indicate that the LMs have not fully adapted to the Z interpretation.
To identify this behavior and quantify it across the data, we consider patterns in the dependency labels predicted for a sentence. In particular, the following condition:
\begin{itemize}
    \item The post-locus cue is not recognized as the main verb;
    \item A comma is placed between the post-locus cue verb and the main verb;
    \item The comma is not followed by a conjunction.
\end{itemize}

\subsection{The Noun/Verb Ambiguity}
\label{appendix_c_nv}

For the Noun/Verb ambiguity, we read the interpretation off the PoS tag predicted by the parser for the locus of ambiguity (\textit{NN, NNS} etc. $\rightarrow$ Noun; \textit{VB, MD} etc. $\rightarrow$ Verb).
Errors in the PoS tags predicted by the parser tend to cause incorrect dependency labels as well; as such, we do not rely on the dependency labels for this ambiguity.

\section{Prompts}
\label{appendix_d}

Tables~\ref{nps-allprompts} through  \ref{nv-allprompts} show the full list of prompts used in our experiments on each ambiguity type.

\begin{table*}[]
\small
\begin{tabular}{lll}
\textsc{No cue} / \textsc{Pre-locus cue} & \textsc{Post-locus cue} / \textsc{Pre\&post-locus cues} & Locus \\ \toprule
The employees understood (that) the contract & The employees understood (that) the contract would   & contract   \\
The mechanic accepted (that) the car         & The mechanic accepted (that) the car looked          & car        \\
The old man recalled (that) the nurse        & The old man recalled (that) the nurse had            & nurse      \\
The traveler heard (that) the clock          & The traveler heard (that) the clock had              & clock      \\
The journalist confirmed (that) the story    & The journalist confirmed (that) the story would      & story      \\
The worker maintained (that) the walls       & The worker maintained (that) the walls fell          & walls      \\
The apprentice forgot (that) the bicycle     & The apprentice forgot (that) the bicycle was         & bicycle    \\
The committee mentioned (that) the issue     & The committee mentioned (that) the issue would       & issue      \\
The army found (that) the supplies           & The army found (that) the supplies saved             & supplies   \\
The umpire warned (that) the spectators      & The umpire warned (that) the spectators would        & spectators \\
The coach discovered (that) the player       & The coach discovered (that) the player tried         & player     \\
The woman noticed (that) the flyer           & The woman noticed (that) the flyer had               & flyer      \\
The tourists saw (that) the palace           & The tourists saw (that) the palace was               & palace     \\
The scientist proved (that) the theory       & The scientist proved (that) the theory could         & theory     \\
The soldiers remembered (that) the town      & The soldiers remembered (that) the town had          & town       \\
The priest recognized (that) two guests      & The priest recognized (that) two guests were         & guests     \\
The reporter revealed (that) the politician  & The reporter revealed (that) the politician received & politician \\
The owners insured (that) the house          & The owners insured (that) the house would            & house      \\
The lawyer established (that) the alibi      & The lawyer established (that) the alibi was          & alibi      \\
The store guaranteed (that) the television   & The store guaranteed (that) the television would     & television \\ \bottomrule
\end{tabular}
\caption{Prompts for NP/S ambiguity (pre-locus cue in parenthesis).}
\label{nps-allprompts}
\end{table*}

\begin{table*}[]
\small
\begin{tabular}{lll}
\textsc{No cue} / \textsc{Pre-locus cue} & \textsc{Post-locus cue} / \textsc{Pre\&post-locus cues} & Locus \\ \toprule
Even though the band left(,) the party            & Even though the band left(,) the party went           & party      \\
In case the executive forgot(,) the assistant     & In case the executive forgot(,) the assistant would   & assistant  \\
Although the maid cleaned(,) the house            & Although the maid cleaned(,) the house was            & house      \\
Because the class failed(,) the exam              & Because the class failed(,) the exam was              & exam       \\
Once the child played(,) the piano                & Once the child played(,) the piano was                & piano      \\
As the couple danced(,) the tango                 & As the couple danced(,) the tango began               & tango      \\
After the kids cheated(,) the teacher             & After the kids cheated(,) the teacher had             & teacher    \\
After the thief attacked(,) the runner            & After the thief attacked(,) the runner was            & runner     \\
Even though the girl phoned(,) the instructor     & Even though the girl phoned(,) the instructor was     & instructor \\
Even though the janitor cleaned(,) the carpet     & Even though the janitor cleaned(,) the carpet was     & carpet     \\
Although the candidates debated(,) the issues     & Although the candidates debated(,) the issues were    & issues     \\
Because the train stopped(,) the traffic          & Because the train stopped(,) the traffic was          & traffic    \\
In case the team lost(,) the tiebreaker           & In case the team lost(,) the tiebreaker was           & tiebreaker \\
After the librarian called(,) the intern          & After the librarian called(,) the intern began        & intern     \\
Even though the army surrendered(,) the territory & Even though the army surrendered(,) the territory was & territory  \\
While the narrator read(,) the story              & While the narrator read(,) the story was              & story      \\
Before the tribe worshipped(,) the idol           & Before the tribe worshipped(,) the idol was           & idol       \\
In case the manager quit(,) the company           & In case the manager quit(,) the company began         & company    \\
As the customer paid(,) the waitress              & As the customer paid(,) the waitress could            & waitress   \\
While the artist painted(,) the furniture         & While the artist painted(,) the furniture was         & furniture \\ \bottomrule
\end{tabular}
\caption{Prompts for NP/Z ambiguity (pre-locus cue in parenthesis).}
\end{table*}

\begin{table*}[]
\footnotesize
\begin{tabular}{p{6cm}p{7cm}l}
\textsc{No cue} / \textsc{Pre-locus cue} & \textsc{Post-locus cue} / \textsc{Pre\&post-locus cues} & Locus \\ \toprule
Mary thinks that the/those pants suit                                             & Mary thinks that the/those pants suit me                                               & suit     \\
The local newspaper reported that the/this warehouse fires                        & The local newspaper reported that the/this warehouse fires numerous                    & fires    \\
We all should have known that the/this metal rings                                & We all should have known that the/this metal rings loudly                              & rings    \\
Susan was extremely surprised that the/this winter bears                          & Susan was extremely surprised that the/this winter bears no                            & bears    \\
A lot of people know that the/a cashier checks                                    & A lot of people know that the/a cashier checks the                                     & checks   \\
Local people are concerned that the/this theater shows                            & Local people are concerned that the/this theater shows lots                            & shows    \\
I know that the/this desert trains                                                & I know that the/this desert trains young                                               & trains   \\
In an old version of that movie, the/a detective cases                            & In an old version of that movie, the/a detective cases the                             & cases    \\
Tom remarked that the/each summer flies                                           & Tom remarked that the/each summer flies by                                             & flies    \\
Despite last year's report, the/this city hopes                                   & Despite last year's report, the/this city hopes that                                   & hopes    \\
Every American knows that the/this government promises                            & Every American knows that the/this government promises it                              & promises \\
Nobody knows if it's true that the/this university fines visitors                 & Nobody knows if it's true that the/this university fines visitors                      & fines    \\
Mrs.~Baker is convinced that the/this school fears                                & Mrs. Baker is convinced that the/this school fears that                                & fears    \\
We just found out that the/this post office packages                              & We just found out that the/this post office packages some                              & packages \\
Some of us weren't aware that the/this church pardons                             & Some of us weren't aware that the/this church pardons very                             & pardons  \\
Nobody seems to complain about the fact that the/this department store buys       & Nobody seems to complain about the fact that the/this department store buys only       & stores   \\
It is no secret that the/this official lies                                       & It is no secret that the/this official lies all                                        & lies     \\
Mrs. Jones is pleased now that she has discovered that the/this greenhouse plants & Mrs. Jones is pleased now that she has discovered that the/this greenhouse plants lots & plants   \\
We should have realized that the/this tractor wrecks                              & We should have realized that the/this tractor wrecks the                               & wrecks   \\
The agency reported that the/this family worries                                  & The agency reported that the/this family worries most                                  & worries  \\
Some people think it's ridiculous that the/this county buses                      & Some people think it's ridiculous that the/this county buses most                      & buses  \\ \bottomrule 
\end{tabular}
\caption{Prompts for Noun/Verb ambiguity, with Verb interpretation.}
\label{nv-allprompts}
\end{table*}

\begin{table*}[]
\footnotesize
\begin{tabular}{p{6cm}p{7cm}l}
\textsc{No cue} / \textsc{Pre-locus cue} & \textsc{Post-locus cue} / \textsc{Pre\&post-locus cues} & Locus \\ \toprule
Mary thinks that the/this pants suit                                               & Mary thinks that the/this pants suit is                                                & suit     \\
The local newspaper reported that the/these warehouse fires                        & The local newspaper reported that the/these warehouse fires harm                       & fires    \\
We all should have known that the/those metal rings                                & We all should have known that the/those metal rings are                                & rings    \\
Susan was extremely surprised that the/those winter bears                          & Susan was extremely surprised that the/those winter bears resemble                     & bears    \\
A lot of people know that the/those cashier checks                                 & A lot of people know that the/those cashier checks are                                 & checks   \\
Local people are concerned that the/many theater shows                             & Local people are concerned that the/many theater shows are                             & shows    \\
I know that the/these desert trains                                                & I know that the/these desert trains are                                                & trains   \\
In an old version of that movie, the/those detective cases                         & In an old version of that movie, the/those detective cases are                         & cases    \\
Tom remarked that the/those summer flies                                           & Tom remarked that the/those summer flies are                                           & flies    \\
Despite last year's report, the/those city hopes                                   & Despite last year's report, the/those city hopes were                                  & hopes    \\
Every American knows that the/these government promises                            & Every American knows that the/these government promises are                            & promises \\
Nobody knows if it's true that the/those university fines                          & Nobody knows if it's true that the/those university fines are                          & fines    \\
Mrs. Baker is convinced that the/these school fears                                & Mrs. Baker is convinced that the/these school fears are                                & fears    \\
We just found out that the/these post office packages                              & We just found out that the/these post office packages are                              & packages \\
Some of us weren't aware that the/these church pardons                             & Some of us weren't aware that the/these church pardons are                             & pardons  \\
Nobody seems to complain about the fact that the/those department store buys       & Nobody seems to complain about the fact that the/those department store buys are       & stores   \\
Mrs. Jones is pleased now that she has discovered that the/those greenhouse plants & Mrs. Jones is pleased now that she has discovered that the/those greenhouse plants are & lies     \\
We should have realized that the/these tractor wrecks                              & We should have realized that the/these tractor wrecks are                              & plants   \\
The agency reported that the/these family worries                                  & The agency reported that the/these family worries are                                  & wrecks   \\
Some people think it's ridiculous that the/those county buses                      & Some people think it's ridiculous that the/those county buses are                      & worries  \\
John quickly learned that the/these hardware store prices                          & John quickly learned that the/these hardware store prices are                          & buses  \\ \bottomrule
\end{tabular}
\caption{Prompts for Noun/Verb ambiguity, with Noun interpretations.}
\label{nv-allprompts-n}
\end{table*}

\end{document}